\documentclass[sigconf]{acmart}
\usepackage{booktabs}
\usepackage{colortbl}
\usepackage{algorithm}  
\usepackage{algorithmic}


\AtBeginDocument{%
  \providecommand\BibTeX{{%
    \normalfont B\kern-0.5em{\scshape i\kern-0.25em b}\kern-0.8em\TeX}}}

\setcopyright{acmcopyright}
\copyrightyear{2022}
\acmYear{2022}
\acmDOI{10.1145/3520304.3529083}

\acmConference[GECCO '22]{The Genetic and Evolutionary Computation Conference 2022}{July 9--13, 2022}{Boston, USA}
\acmBooktitle{GECCO '22: The Genetic and Evolutionary Computation Conference 2022, July 9--13, 2022, Boston, USA}
\acmPrice{15.00}
\acmISBN{978-1-4503-9268-6/22/07}

\acmDOI{10.1145/3520304.3529083}
\acmISBN{978-1-4503-9268-6/22/07} % To be updated after completing copyright process
\acmConference[GECCO '22]{The Genetic and Evolutionary Computation Conference 2022}{July 9--13, 2022}{Boston, USA}
\acmYear{2022}
\copyrightyear{2022}

\begin{document}

\title{Dynamic Multi-objective Ensemble of Acquisition Functions in Batch Bayesian Optimization}

% \author{Anonymous Authors}

\author{Jixiang Chen}
\authornotemark[1]
\orcid{0000-0001-9941-8324}
\author{Fu Luo}
\orcid{0000-0002-3161-6348}
\authornote{Both authors contributed equally to this research.}
\affiliation{%
  \institution{School of System Design and Intelligent Manufacturing,\\ Southern University of Science and Technology}
  \streetaddress{No. 1088, Xueyuan Avenue,Nanshan District}
  \city{Shenzhen}
%   \state{Guangdong}
  \country{China}
  \postcode{518055}
}
\email{{11911816,11910310}@mail.sustech.edu.cn}

\author{Zhenkun Wang}
\authornote{Corresponding author.}
\orcid{0000-0003-1152-6780}
\affiliation{%
  \institution{School of System Design and Intelligent Manufacturing,\\ Southern University of Science and Technology}
  \streetaddress{1 Th{\o}rv{\"a}ld Circle}
  \city{Shenzhen}
  \country{China}}
\email{wangzk3@sustech.edu.cn}

% \renewcommand{\shortauthors}{Jixiang Chen, Fu Luo and Zhenkun Wang}

\begin{abstract}
    Bayesian optimization (BO) is a typical approach to solve expensive optimization problems. In each iteration of BO, a Gaussian process (GP) model is trained using the previously evaluated solutions; then next candidate solutions for expensive evaluation are recommended by maximizing a cheaply-evaluated acquisition function on the trained surrogate model. The acquisition function plays a crucial role in the optimization process. However, each acquisition function has its own strengths and weaknesses, and no single acquisition function can consistently outperform the others on all kinds of problems. To better leverage the advantages of different acquisition functions, we propose a new method for batch BO. In each iteration, three acquisition functions are dynamically selected from a set based on their current and historical performance to form a multi-objective optimization problem (MOP). Using an evolutionary multi-objective algorithm to optimize such a MOP, a set of non-dominated solutions can be obtained. To select batch candidate solutions, we rank these non-dominated solutions into several layers according to their relative performance on the three acquisition functions. The empirical results show that the proposed method is competitive with the state-of-the-art methods on different problems.
\end{abstract}






\maketitle

\section{Introduction}

Many real-world applications are black-box function optimization problems. They have no closed-form expression or derivative information and are often expensive to evaluate. Bayesian optimization (BO) is an efficient method to deal with these problems~\cite{human}. BO methods leverage a Gaussian Process~\cite{GP} (GP) to approximate the original black-box function with a much cheaper evaluation cost. In each iteration, the next solution for expensive evaluation is recommended by maximizing an acquisition function that balances exploitation and exploration by leveraging GP posterior information. 

It is possible to evaluate different solutions in parallel at one iteration, which leads to the study of batch BO (BBO) algorithms. As stated in~\cite{Po}, no single acquisition can consistently outperform the others on all kinds of problems. Subsequently, a group of BBO algorithms adopts multiple classical acquisition functions to form a multi-objective optimization problem (MOP). For example,~\cite{MACE} takes three different acquisition functions to form a MOP, and the obtained non-dominated solutions can balance between the three acquisition functions. Although MOP-based algorithms can avoid the inefficiencies raised by a single acquisition function's bias, they still have shortcomings. 1) They employ only a few acquisition functions and are not sufficiently capable of covering many different types of problems. 2) They use fixed acquisition functions whose performance may vary drastically during optimization. 3) They have rarely studied the selection strategy. The batch solutions are randomly selected from the obtained non-dominated solutions.

To fill these gaps, we propose to adopt dynamic multi-objective ensemble of acquisition functions (DMEA) for the BBO algorithm. In each iteration, DMEA dynamically selects three acquisition functions from a set to form a MOP using a heuristic strategy based on their current and historical performance. The MOP is always constructed with the most appropriate acquisition functions at the current optimization stage. By implementing an evolutionary multi-objective optimization algorithm to optimize the MOP, the acquired non-dominated solutions reflect the best trade-off locations among the three well-performed acquisition functions. To better select a batch of solutions in the non-dominated set, we propose a preferred select strategy to rank these solutions into different layers according to their relative performance in the three acquisition functions. We empirically show that the proposed algorithm gives competitive performance compared with different state-of-the-art methods for BBO on different benchmark functions.

\section{Proposed BBO Algorithm: DMEA}

\begin{algorithm}
\caption{DMEA}
\label{PMACE}
\begin{algorithmic}[1]
  \REQUIRE
  Initial data $S$, number of iterations $N_{iter}$, batch size $k$, candidate $q$ acquisition functions $\alpha^1(\mathbf{x}), \alpha^2(\mathbf{x}), \ldots, \alpha^q(\mathbf{x})$, hyperparameter $\eta$
  \STATE $S_0, X_0 \leftarrow$ Select $k$ data pairs with the smallest $f(\mathbf{x})$ values from $S$
  \STATE $S_h \leftarrow S - S_0 $ 
  \STATE GP $\leftarrow$ Train the initial GP model using $S_h$
  \FOR{$i = 1 \textbf{ to } N_{iter}$}
  \STATE Calculate the quality indicator  $hq(\mathbf{x})$ for each $\mathbf{x} \in X_{i-1}$ using equation (\ref{phih})
  \FOR{$j = 1 \textbf{ to } q$}
  \STATE $p_j \leftarrow 0$
 \STATE  $g^j \leftarrow$ Apply LP method to $\alpha(\mathbf{x})^j$ 
  \FOR{$\mathbf{x} \in X_{i-1}$}
    \STATE Calculate $\varphi_\alpha^j(\mathbf{x})$ by equation (\ref{phia})
  \ENDFOR
  \STATE Calculate $p_j$ and $P_j$ by equations (\ref{cp}) and (\ref{pp}) respectively

  \ENDFOR
  \STATE $S_h \leftarrow S_h \cup S_{i-1}$
  \STATE Train the GP model using $S_h$
  \STATE $\alpha_1'(\mathbf{x}),\ldots,\alpha_3'(\mathbf{x}) \leftarrow $ Select three acquisition functions from $q$ candidates with smallest three values of corresponding $P_j$ 
  \STATE $f_1(\mathbf{x}),f_2(\mathbf{x}), f_3(\mathbf{x}) \leftarrow$ Multiply negative one with $\alpha_1'(\mathbf{x}),\ldots,\alpha_3'(\mathbf{x}) $ to form the standard MOP
    \STATE $\boldsymbol{T} \leftarrow $ Calculate the confidence vector
    \STATE $\Omega_{PF}, \Omega_{PS} \leftarrow $ Implement NSGA-II to optimize such a MOP
    \STATE $X_c \leftarrow $ Select $m$ solutions in $\Omega_{PS}$ corresponding to extreme solutions in $\Omega_{PF}$
    \STATE $X_r \leftarrow $ Apply preferred select strategy to get remaining $k-m$ solutions
    \STATE $X_c \leftarrow X_c \cup X_r$ 
    \STATE $f(\mathbf{x}_1),\ldots,f(\mathbf{x}_k) \leftarrow$ expensively evaluate $X_c = \{\mathbf{x}_1,\ldots, \mathbf{x}_k\}$ 
    \STATE $S_i \leftarrow \{(\mathbf{x}_1, f(\mathbf{x}_1)),\ldots,(\mathbf{x}_k, f(\mathbf{x}_k)) \}$
    \ENDFOR
\ENSURE{Best $f(\mathbf{x}^*)$ expensively evaluated during iterations}
\end{algorithmic}
\end{algorithm}

\subsection{DMEA}
The details of the proposed \textit{DMEA} are introduced in Algorithm~\ref{PMACE}. In the $i$-th iteration, current data is divided into two parts. The first part is the data recommended in the $(i-1)$-th iteration, i.e., $S_{i-1} = \{(\mathbf{x}_u, f(\mathbf{x}_u))\}_{u = 1}^{k}$. The other part denoted by $S_h =  S - S_{i-1}$ is the historical data used to get $S_{i-1}$. This division helps us exploit information from the last batch solutions to assess the quality of each acquisition function. We evaluate each acquisition function by comparing the solutions it recommends in $(i-1)$-th iteration with the historical solutions. Considering the computational costs in reconstructing GP models, we use the entire last batch rather than one-by-one from them for acquisition function evaluation. $S_0$ is initialized with the $k$ solutions having the smallest function values. The initial GP model is trained with $S-S_0$ (as shown in Lines 1-3 of Algorithm~\ref{PMACE}).

At the beginning of the $i$-th iteration, we measure the quality of the solutions in $S_{i-1}$ by comparing them with historical data. As shown in Line 5, we employ equation (\ref{phih}) to calculate the quality indicator for each $\mathbf{x}\in X_{i-1}$. 
\begin{equation}
    hq(\mathbf{x}) = \left\{\begin{matrix} 
  1, \quad num(\mathbf{x}) \le 3 \\  
  0, \quad num(\mathbf{x}) > 3 
\end{matrix}\right., \label{phih}
\end{equation} 
where $num(\mathbf{x})$ denotes the number of solutions in $S_{h}$ whose function value is better than $\mathbf{x}$. The currently recommended solutions usually outperform most historical solutions. They are regarded as high-quality ones if they are only beaten by a small number of historical solutions. We set a fixed threshold of 3 here, according to experimental tests.

Furthermore, to judge the performance of an acquisition function, we need to identify whether it recommends a good solution. An intuitive way is to identify a threshold of the acquisition function value because a larger value corresponds to more recommendations for a solution. 
To get the threshold, it is natural to first find the optimal of an acquisition function, then the sub-optimal, and repeat the process until finding a satisfactory value. A popular BBO method called Local Penalization (LP) is suitable for finding the threshold. It behaves similarly by finding the optimal solution, then penalizing the region around it and repeating the process to find sub-optimal solutions. So we implement LP to recommend solutions to help us find the threshold. To decide the number of solutions recommended by one acquisition function using LP, we consider that three acquisition functions recommend each batch of solutions in the proposed method. The number is k/3 for each acquisition function on average. As the solution could be recommended by more than one acquisition function, we adopt $\lfloor k/2 \rfloor$ instead of k/3. So for the $j$-th acquisition function $\alpha(\mathbf{x})^j$, we implement LP to recommend $\left \lfloor k/2 \right \rfloor$ solutions, denoted by $X_{\alpha}^j = \{\mathbf{x}_u\}_{u = 1}^{\left \lfloor k/2 \right \rfloor}$. And the smallest value of $\alpha(\mathbf{x})^j, \mathbf{x}\in X_{\alpha}^j$ is used as the threshold, denoted by $g^j$ (as shown in Lines 7-10 of Algorithm~\ref{PMACE}). For $\mathbf{x} \in X_{i-1}$, if $\alpha(\mathbf{x})^j < g^j$, the acquisition function is regarded as not recommending the solution, otherwise it recommends the solution, i.e.,
\begin{equation}
    \varphi_\alpha^j(\mathbf{x}) = \left\{\begin{matrix} 
  1, \quad \alpha^j(\mathbf{x}) \ge g^j \\  
  0, \quad \alpha^j(\mathbf{x}) < g^j.
\end{matrix}\right. \label{phia}
\end{equation}

So we can combine $hq(\mathbf{x})$ and  $\varphi^j_\alpha(\mathbf{x})$ to define \textit{Cumulative Penalty} and \textit{Recent Penalty} (denoted by $P_j$ and $p_j$ respectively) to evaluate each $\alpha(\mathbf{x})^j$ (as shown in Line 12 of Algorithm~\ref{PMACE}). $p_j$ only uses the information of the previous iteration, while $P_j$ contains all the historical information as it is updated by $p_j$ iteratively. Intuitively, a good candidate acquisition function should only recommend high-quality solutions. $\alpha(\mathbf{x})^j$ can have four situations for each $\mathbf{x} \in X_{i-1}$. (1) If it does not recommend solutions with $hq(\mathbf{x})=0$, neither penalty nor reward would be calculated for $p_j$. (2) If it misses solutions with $hq(\mathbf{x})=1$ or (3) it recommends solutions with $hq(\mathbf{x})=0$, it will receive a penalty for this misbehavior. (4) If it recommends solutions with $hq(\mathbf{x})=1$, it could receive a reward. The amount of penalty and reward should consider the amount of absolute deviation from current optimum. With these considerations, $p_j$ is calculated as:
\begin{equation}
\begin{aligned}
        p_j &= \sum_{u = 1}^k\Big(\left | hq(\mathbf{x}_u) - \varphi_{\alpha}^j(\mathbf{x}_u) \right |\cdot\left | f(\mathbf{x}_u) - f(\mathbf{x}^*)\right |\\  &+ hq(\mathbf{x}_u)\cdot\varphi_{\alpha}^j(\mathbf{x}_u)\cdot\left ( f(\mathbf{x}_u) - f(\mathbf{x}^*)\right )\Big), \label{cp}
\end{aligned}
\end{equation}
where $\mathbf{x}_u \in X_{i-1}$ , and $f(\mathbf{x}^*)$ is the best value for each solution in $S_h$. Then $P_j$ is updated as:
\begin{equation}
    P_j\leftarrow  \eta P_j + p_j, \label{pp}
\end{equation}
where $\eta$ is a hyperparameter to control the influence of historical performance. 

Then we update $S_h$ with $S_{i-1}$ and train the GP model with the updated $S_h$ (as shown in Lines 14-15 of Algorithm~\ref{PMACE}). For dynamic selection, We choose the three acquisition functions $\{\alpha(\mathbf{x})_u'\}_{u = 1}^3$ with the smallest three $P_j$ which give the best performance at current optimization stage. Problems for the standard MOP are denoted by $f(\mathbf{x})_u = -\alpha(\mathbf{x})_u', u=1,2,3$. NSGA-II~\cite{NSGA} is then applied to optimize such a MOP to get the non-dominated solution set in both objective space and the design space, denoted by $\Omega_{PF}$ and $\Omega_{PS}$ respectively. The corresponding $P_j$, denoted by $\{P'_u\}_{u=1}^3$, are used to form the \textit{confidence vector}, denoted by $\boldsymbol{T} = \begin{bmatrix} t_1, t_2, t_3 \end{bmatrix}^T = \begin{bmatrix} P'_1, P'_2, P'_3 \end{bmatrix}^T$. $t_j$ represents the relative performance of $\alpha(\mathbf{x})'_j$ among the three (as shown in Lines 16-19 of Algorithm~\ref{PMACE}).

To select solutions in the non-dominated set, we propose to adopt the preferred select strategy that exploits the confidence vector to rank these solutions into different layers. Before that, we first select extreme solutions in $\Omega_{PF}$ corresponding to each smallest single objective value. Because some objectives could be non-conflicting, their smallest objectives value could correspond to the same solution. We first include these $m$ extreme solutions, denoted by $X_c$. Then the preferred select strategy is applied to select the remaining $(k-m)$ solutions (as shown in Line 20-21 of Algorithm~\ref{PMACE}).

\subsection{Preferred select strategy}
\label{SecPSS}
The implementation procedure is as follows:

Firstly, we select the threshold for preferred recommendation. For $s$ solutions in $\Omega_{PF} = \{\mathbf{f}(\mathbf{x})_i\}_{i=1}^s$, we set the $\lfloor \frac{s}{5} \rfloor$ smallest value of each objective $f_{j}(\mathbf{x})$ as the threshold $b_j$, which is a practical value in test. $b_j$ selects a relative small region to reflect preferred recommendation by $\alpha(\mathbf{x})'_j$.

Then we divide solutions into different layers. For every two objectives of $\mathbf{f}(\mathbf{x})_i$, e.g., $f_u(\mathbf{x})_i, f_v(\mathbf{x})_i$. 
Suppose $t_u < t_v$, this reflects that $\alpha(\mathbf{x})_u'$ gives better performance than $\alpha(\mathbf{x})_v'$. Meantime, if $f_u(\mathbf{x})_i \le b_u$, the solution is regarded as to be preferred recommended by $\alpha(\mathbf{x})_u'$ and the \textit{confident level} $c_i$ will plus one. Thus we can divide the non-dominated set into different layers with different $c_i$ ranging from 0 to 3. We define that the solutions with confident level $i$ is in $i$-th preferred layer, denoted by $C_i$.

Finally, we select solutions from different preferred layers. Because a higher confident level of a solution means that it is more recommended by the most suitable acquisition function, we select the first two non-empty layers from high confident to low confident levels, denoted by $C_1'$ and $C_2'$ respectively. To select the solutions, $\lceil \frac{2}{3}(k-m) \rceil$ solutions are randomly sampled from $C_1'$, the other  solutions are randomly sampled from $C_2'$.

\begin{table*}[t]
\centering
\caption{Optimization results of the tested algorithms with batch sizes of $k=\{4,10\}$. The best algorithm is in a grey background; the statistically equivalent best algorithm is in blue.}
\label{tab:compare result table}
 \resizebox{0.72\textwidth}{!}{%
\begin{tabular}{c|
>{\columncolor[HTML]{FFFFFF}}c c|
>{\columncolor[HTML]{FFFFFF}}c c|
>{\columncolor[HTML]{FFFFFF}}c c|
>{\columncolor[HTML]{FFFFFF}}c c}
\toprule[0.6pt]
\cellcolor[HTML]{FFFFFF}\textbf{Algorithm} & \multicolumn{2}{c|}{\cellcolor[HTML]{FFFFFF}\textbf{SixHumpCamel (d=2)}}                         & \multicolumn{2}{c|}{\cellcolor[HTML]{FFFFFF}\textbf{Eggholder (d=2)}}       & \multicolumn{2}{c|}{\cellcolor[HTML]{FFFFFF}\textbf{Branin (d=2)}}        & \multicolumn{2}{c}{\cellcolor[HTML]{FFFFFF}\textbf{Ackley2 (d=2)}}          \\
\cellcolor[HTML]{FFFFFF}                &\cellcolor[HTML]{FFFFFF} k=4      & \cellcolor[HTML]{FFFFFF}k=10                            &\cellcolor[HTML]{FFFFFF} k=4                               & \cellcolor[HTML]{FFFFFF}k=10      &\cellcolor[HTML]{FFFFFF} k=4                              & \cellcolor[HTML]{FFFFFF}k=10     & {\color[HTML]{000000} k=4}        & \cellcolor[HTML]{FFFFFF}k=10      \\ \midrule[0.6pt]
\cellcolor[HTML]{FFFFFF}DMEA            & \cellcolor[HTML]{C0C0C0}7.290e-5 & \cellcolor[HTML]{FFFFFF}{\color[HTML]{3531FF} 7.201e-5} & \cellcolor[HTML]{C0C0C0}34.19     & {\color[HTML]{3531FF}22.61}     & \cellcolor[HTML]{C0C0C0}5.757e-8 & \cellcolor[HTML]{C0C0C0}2.314e-8 & {\color[HTML]{3531FF} 0.2308}     & {\color[HTML]{3531FF} 0.1693}     \\
\cellcolor[HTML]{FFFFFF}MACE            & 8.451e-5                         & \cellcolor[HTML]{FFFFFF}7.802e-5                        & 79.29                             & {\color[HTML]{3531FF} 35.12}      & 3.732e-6                         & 2.434e-6                         & 0.5871                            & 0.4028                            \\
\cellcolor[HTML]{FFFFFF}EI-LP           & 9.921e-5                         & \cellcolor[HTML]{FFFFFF}9.016e-5                        & {\color[HTML]{3531FF} 35.72}      & \cellcolor[HTML]{C0C0C0}14.60     & 2.068e-4                         & 1.168e-4                         & \cellcolor[HTML]{C0C0C0}0.1687    & \cellcolor[HTML]{C0C0C0}0.0571    \\
\cellcolor[HTML]{FFFFFF}qEI             & 7.355e-5                         & \cellcolor[HTML]{FFFFFF}7.557e-5                        & {\color[HTML]{3531FF} 46.25}      & {\color[HTML]{3531FF} 17.68}      & 1.0261e-5                        & 1.263e-5                         & 0.7282                            & 0.4581                            \\
\cellcolor[HTML]{FFFFFF}$\epsilon$S-PF          & {\color[HTML]{3531FF} 7.425e-5}  & \cellcolor[HTML]{C0C0C0}7.182e-5                        & 186.37                            & 129.04                            & 8.545e-7                         & {\color[HTML]{3531FF} 6.248e-7}  & 0.3798                            & 0.2102                            \\ \midrule[0.6pt]
\cellcolor[HTML]{FFFFFF}\textbf{Algorithm} & \multicolumn{2}{c|}{\cellcolor[HTML]{FFFFFF}\textbf{Rosenbrock2 (d=2)}}                          & \multicolumn{2}{c|}{\cellcolor[HTML]{FFFFFF}\textbf{BraninForrester (d=2)}} & \multicolumn{2}{c|}{\cellcolor[HTML]{FFFFFF}\textbf{Alpine1 (d=5)}}       & \multicolumn{2}{c}{\cellcolor[HTML]{FFFFFF}\textbf{Hartmann6 (d=6)}}        \\
                                        & \cellcolor[HTML]{FFFFFF}k=4                              & \cellcolor[HTML]{FFFFFF}k=10                            &\cellcolor[HTML]{FFFFFF} k=4                               & \cellcolor[HTML]{FFFFFF}k=10      &\cellcolor[HTML]{FFFFFF} k=4                              & \cellcolor[HTML]{FFFFFF}k=10     &\cellcolor[HTML]{FFFFFF} k=4                               & \cellcolor[HTML]{FFFFFF}k=10      \\ \midrule[0.6pt]
\cellcolor[HTML]{FFFFFF}DMEA            & \cellcolor[HTML]{C0C0C0}1.359e-3 & {\color[HTML]{3531FF} 6.206e-4}                         & \cellcolor[HTML]{C0C0C0}3.465e-6  & \cellcolor[HTML]{C0C0C0}3.439e-6  & {\color[HTML]{3531FF} 0.1894}    & {\color[HTML]{3531FF} 0.1228}    & \cellcolor[HTML]{C0C0C0}0.0476907 & \cellcolor[HTML]{C0C0C0}0.0420372 \\
\cellcolor[HTML]{FFFFFF}MACE            & {\color[HTML]{3531FF} 2.322e-3}  & \cellcolor[HTML]{C0C0C0}2.933e-4                        & 4.524e-6                          & 4.950e-6                          & {\color[HTML]{3531FF} 0.2186}    & \cellcolor[HTML]{C0C0C0}0.03245  & {\color[HTML]{3531FF} 0.0477051}  & {\color[HTML]{3531FF} 0.0476925}  \\
\cellcolor[HTML]{FFFFFF}EI-LP           & 8.558e-3                         & 2.454e-3                                                & 4.457e-4                          & 1.296e-4                          & 0.6684                           & 0.2267                           & 0.0596396                         & {\color[HTML]{3531FF} 0.0420957}  \\
\cellcolor[HTML]{FFFFFF}qEI             & 9.500e-3                         & 7.355e-3                                                & 3.141e-5                          & 2.362e-5                          & 0.9665                           & 0.7100                           & 0.1106420                         & 0.1029673                         \\
\cellcolor[HTML]{FFFFFF}$\epsilon$S-PF          & 2.423e-1                         & 1.936e-1                                                & 9.562e-6                          & 4.192e-6                          & \cellcolor[HTML]{C0C0C0}0.1424   & {\color[HTML]{3531FF} 0.0529}    & 0.0538524                         & {\color[HTML]{3531FF} 0.0536522}  \\ \bottomrule[0.6pt]
\end{tabular}%
}
\end{table*}

% \begin{figure*}[htbp]
%     \centering
%     \includegraphics[width=\textwidth]{all_plot_batch_4_and_10.pdf}
%     \caption{Optimization results of the tested algorithms on eight benchmark functions.  }
%     \label{all_plot}
% \end{figure*} 

\section{Experiment}
The proposed DMEA
$\footnote{https://github.com/JixiangChen-Jimmy/DMEA}$
algorithm is tested on eight benchmark functions. The hyperparameter $\eta$ is empirically suggested to be set as 0. Four state-of-the-art BBO methods are compared, including the local penalization method with EI acquisition function
(EI-LP)~\cite{LP}$\footnote{https://github.com/SheffieldML/GPyOpt}$,
the batch expected improvement
(qEI)~\cite{qEI}$\footnote{\label{note1}https://github.com/georgedeath/eshotgun}$,
the multi-objective acquisition ensemble
(MACE)~\cite{MACE}$\footnote{https://github.com/Alaya-in-Matrix/pyMACE}$, and the  $\epsilon$-greedy Batch Bayesian Optimisation  with $\epsilon=0.1$ ($\epsilon$S-PF)~\cite{eshotgun}$\footnote{\label{note1}https://github.com/georgedeath/eshotgun}$. To optimize the MOP in each iteration, we implement the NSGA-II~\cite{NSGA} algorithm. For the GP model, the ARD Matern 5/2 kernel is applied. The hyperparameters of the GP model are optimized via the maximization of the log likelihood.
% The candidate acquisition functions are set similar to MACE.

 For each algorithm, we use Latin Hypercube Sampling~\cite{LHS} to get the initial data. The number of initial data is $N_{init} = 11d-1$ where $d$ is the number of dimension of the benchmark function. Referring to the setting of \cite{MACE}, we set the number of iterations to $N_{iter} = 45$. And the batch size of $k=\{4,10\}$ were tested. The total number of function evaluations is $N_{init} + k\cdot N_{iter}$.

\subsection{Candidate acquisition functions}
The candidate acquisition functions consist of one EI with a jitter of $10^{-3}$, one PI with a jitter of $10^{-3}$, and five LCB with different hyperparameter settings:
\begin{equation}
(\nu, \delta)\in\{(0.5,0.5), (0.5,0.05), (5,0.1), (10,0.1), (30,0.1)\},
\end{equation}
where the hyperparameter $\kappa$ of LCB is calculated as:
\begin{equation}
    \begin{aligned}
\kappa &=\sqrt{\nu \tau_{i}}, \\
\tau_{i} &=2 \log \left(i^{d / 2+2} \pi^{2} / 3 \delta\right).
\end{aligned}
\end{equation}

We choose the acquisition functions for the following reasons: We adopt the same settings as MACE, which uses three acquisition functions to form the MOP. The selected acquisition functions are popular in BO, such as PI, EI, and LCB. For LCB, its performance is highly influenced by $\kappa$, so we set several LCB variants with different hyperparameters as the candidate acquisition functions. Besides, the efficiency of MOEAs is also our motivation for selecting three acquisition functions to form the MOP.

\subsection{Result and analysis}
Regret is defined as the positive difference between the evaluated and true optimum.
Table \ref{tab:compare result table} shows the regret of the eight benchmark functions optimized by these five algorithms over 20 runs.
The unilateral Wilcoxon signed-rank test~\cite{wilcoxon} is implemented to find statistically equivalent best algorithms. 

As shown in Table \ref{tab:compare result table}, the number of times DMEA achieves the best performance is six when $k=4$, and the number is three when $k=10$. It can achieve the best or statistically equivalent best performance on all eight functions for both $k=4,10$. 

\begin{figure}[H]
    \centering
    \includegraphics[width=3.2in]{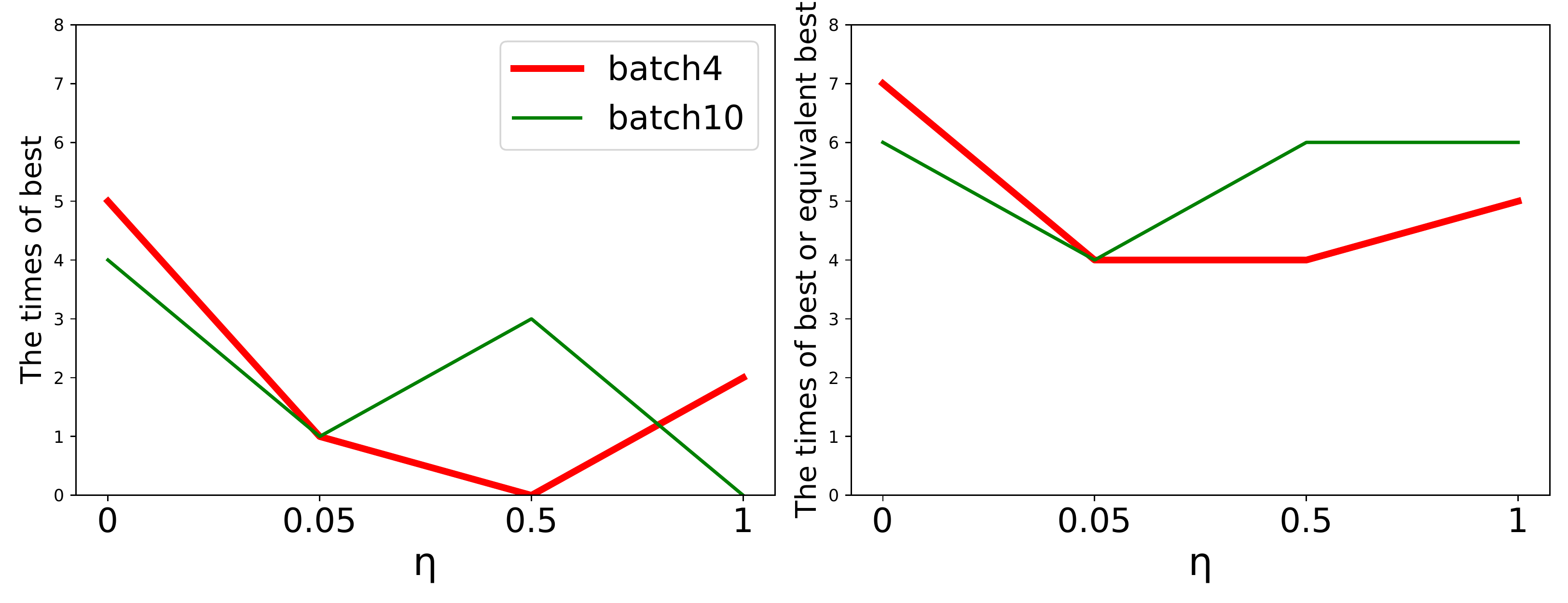}
    \caption{
    Performance of the proposed algorithm with different $\eta$ settings on the eight benchmark functions.
    % The number of times that the performance of the proposed algorithm with different $\eta$ is the best or statistically equivalent best on the eight benchmark functions, for $\eta=\{ 0,0.05,0.5,1\} $ $k=\{4,10\}.$
    }
    \label{eta_compare}
\end{figure}

As shown in Figure \ref{eta_compare}, different settings of hyperparameter $\eta$ are tested. We find that the result of extreme small $\eta$ ($\eta=0$) outperforms that of larger $\eta$ ($\eta=0.05,0.5,1$) on different batch sizes.
In the early optimization stage, exploration behavior should be encouraged because information about many regions in the design space remains unknown. But exploration behavior does not guarantee good solutions, which leads to an increase in the Cumulative Penalty of these strategies quickly. Setting a small $\eta$ could encourage these strategies by decreasing the weight of cumulative behavior. Thus, swift tunning for a good combinational strategy of different optimization stages could be achieved. $\eta=0$ is the extreme case that only uses the previous iteration's information. Therefore, the setting of $\eta=0$ is suggested in the proposed algorithm. The full optimization results of the proposed algorithm with different $\eta$ are shown in the supplementary material.

Besides, we perform the ablation study of the preferred select strategy. The experiment shows the effectiveness of this strategy as it is always better than the random search method except for one benchmark with k=10. Details of the ablation study are included in the supplementary material.

% Different settings to hyperparameter $\eta$ were also tested. We found that the result of small $\eta$ ($\eta=0,0.05$) outperformed that of large $\eta$ ($\eta=0.5,1$). Intuitively, a small $\eta$ is more 'greedy' using more recent penalty information, thus swift tunning for the good combinational strategy could be achieved. $\eta=0$ is the extreme case of only using the previous iteration's information. A large $\eta$ could lead to slow tunning which is mislead by information unsuitable for current optimization stage, thus may lead to poor performance. So a small $\eta$ ($\eta=0,0.05$) is suggested to be applied in proposed method.

\section{Conclusion and Future work}
In this paper, we have proposed a novel method DMEA for BBO. The proposed method exploits three acquisition functions with good performance among the candidates. In each iteration, the three selected ones were used to form a MOP. The non-dominated solutions that reflect different trade-offs among the three different acquisition functions were obtained by optimizing such a MOP. To select the batch solutions from the non-dominated set, we have utilized the information of relative performance between the three acquisition functions to divide solutions into different layers. The empirical results have shown that the proposed method is competitive with the state-of-the-art methods on eight benchmark problems. 

Our future research focuses on three aspects: (1) Further study of hyperparameters in DMEA. (2) Thorough study of candidate acquisition functions that suit DMEA better. (3) Better methods to assess the acquisition functions.

\begin{acks}
This paper is founded by National Natural Science Foundation of China (Grant No:62106096) and Special Funds for the Cultivation of Guangdong College Students' Scientific and Technological Innovation.("Climbing Program" Special Funds) pdjh2022c0093

\end{acks}

\bibliographystyle{ACM-Reference-Format}
\bibliography{reference}

% \appendix









\end{document}